\documentclass[11pt]{article}

\usepackage[preprint]{acl}

\usepackage{times}
\usepackage{latexsym}

\usepackage[T1]{fontenc}

\usepackage[utf8]{inputenc}

\usepackage{microtype}

\usepackage{inconsolata}

\usepackage{graphicx}

\usepackage{tikz}
\usetikzlibrary{shapes.geometric, arrows.meta, positioning, fit, backgrounds}
\usepackage{adjustbox}
\usepackage{amsmath}
\usepackage{amssymb}
\usepackage[most]{tcolorbox}

\newtcolorbox{promptbox}{
  colback=gray!5,
  colframe=gray!60,
  boxrule=0.5pt,
  arc=2pt,
  left=6pt,
  right=6pt,
  top=6pt,
  bottom=6pt
}

\usepackage{bbm}

\usepackage{amsmath}
\usepackage{amssymb}
\usepackage{amsfonts}
\usepackage{amsthm}
\usepackage{mathrsfs}
\usepackage{mathabx}
\usepackage{bm}

\usepackage{booktabs}
\usepackage{multirow}
\usepackage{tabularx}
\usepackage{arydshln}
\usepackage{makecell}
\usepackage{colortbl}
\usepackage{siunitx}

\usepackage{graphicx}
\usepackage{svg}
\usepackage{caption}
\usepackage{subcaption}
\usepackage{wrapfig}
\usepackage{floatflt}
\usepackage{algorithm}
\usepackage{algorithmicx}
\usepackage{algpseudocode}

\usepackage{microtype}
\usepackage{nicefrac}
\usepackage{enumitem}
\usepackage{url}
\usepackage{xurl}
\usepackage{authblk}
\usepackage{lipsum}

\usepackage{soul}

\usepackage[most]{tcolorbox}

\usepackage{fontawesome5}
\usepackage{CJKutf8}
\usepackage{hyperref}
\usepackage{bbding}
\usepackage{pifont}
\usepackage{wasysym}
\usepackage{tabularx}

\newtheorem{proposition}{Proposition}
\newcommand{\method}{RL-Guided Sampling}

\title{Small RL Controller, Large Language Model: RL-Guided Adaptive Sampling for Test-Time Scaling}
\author{
    Runpeng Dai$^{1}$\quad
    Tong Zheng$^{2}$ \quad
    Rui Liu$^{2}$ \quad
    Chengsong Huang$^{3}$ \quad
    Hongtu Zhu$^{1\dag}$ \\
    $^{1}$University of North Carolina at Chapel Hill \quad
    $^{2}$University of Maryland, College Park \\
    $^{3}$Washington University in St. Louis \\

    \texttt{\{runpeng, htzhu\}@unc.edu} \\
    {\footnotesize \textbf{Code:} \href{https://github.com/RunpengDai/RL-Guided-Adaptive-Sampling}{https://github.com/RunpengDai/RL-Guided-Adaptive-Sampling}}
}

\begin{document}

\maketitle

\begin{abstract}
Test-time scaling improves the reasoning performance of large language models but incurs substantial cost in both total computation and latency. Existing adaptive sampling methods partially mitigate this issue by dynamically deciding when to stop sampling, yet they typically rely on heuristic rules or rely on distribution assumptions. In this work, we formulate adaptive sampling as a Markov decision process (MDP). We train a lightweight sampling controller with reinforcement learning (RL) to jointly balance answer correctness, latency, and computation cost. At each round, the controller decides to stop sampling or to acquire additional samples. Our method is lightweight which only relies on statistics of final answers, and can be trained and deployed on CPU. We further show that the resulting framework admits an interpretation as the Lagrangian relaxation of a constrained optimization problem with explicit budget constraints. Experiments against strong baselines such as ASC and ESC show that our method achieves improved trade-offs among answer correctness, sampling rounds, and total samples required. 
\end{abstract}


\section{Introduction}
Test-time scaling \citep{snell2024scaling, zhang2025survey} has emerged as an effective way to improve the reasoning performance of large language models (LLMs) without additional training. Methods such as self-consistency \citep{wang2022self}, tree-of-thoughts \citep{yao2023tree}, and Best-of-$N$ sampling \citep{nakano2021webgpt, huang2025best} improve final answer quality by allocating more inference-time computation. However, these gains come with a clear downside: increased inference cost. As a result, effectively allocating inference-time computation to balance cost and performance is a critical challenge. 


%

A growing line of work seeks to reduce the cost of test-time scaling through adaptive sampling. Early methods, such as Adaptive Self-Consistency (ASC) \citep{aggarwal2023let}, utilize the posterior answer distribution to determine whether additional samples are needed. Subsequent works build on this by incorporating semantic signals \citep{wan2025reasoning} or altering the prior distribution \citep{komiyama2026best}. Meanwhile, Early-Stopping Self-Consistency (ESC) \citep{li2024escape} reduces latency by shifting to a parallel execution strategy. A concurrent line of research attempts to alter the reasoning process itself. At inference time, some methods trigger early stopping via signals such as confidence \citep{Fu2025DeepTW}, probing \citep{Mao2025EarlySC, zheng2026parallel}, or convergence dynamics \citep{liu2025answer, zhang2025alphaone}. Alternatively, other works aim to increase reasoning efficiency during training time, leveraging either supervised fine-tuning \citep{xia2025tokenskip, munkhbat2025self} or reinforcement learning \citep{aggarwal2025l1}.


\begin{figure*}[t!]
  \centering
  \includegraphics[width=\linewidth]{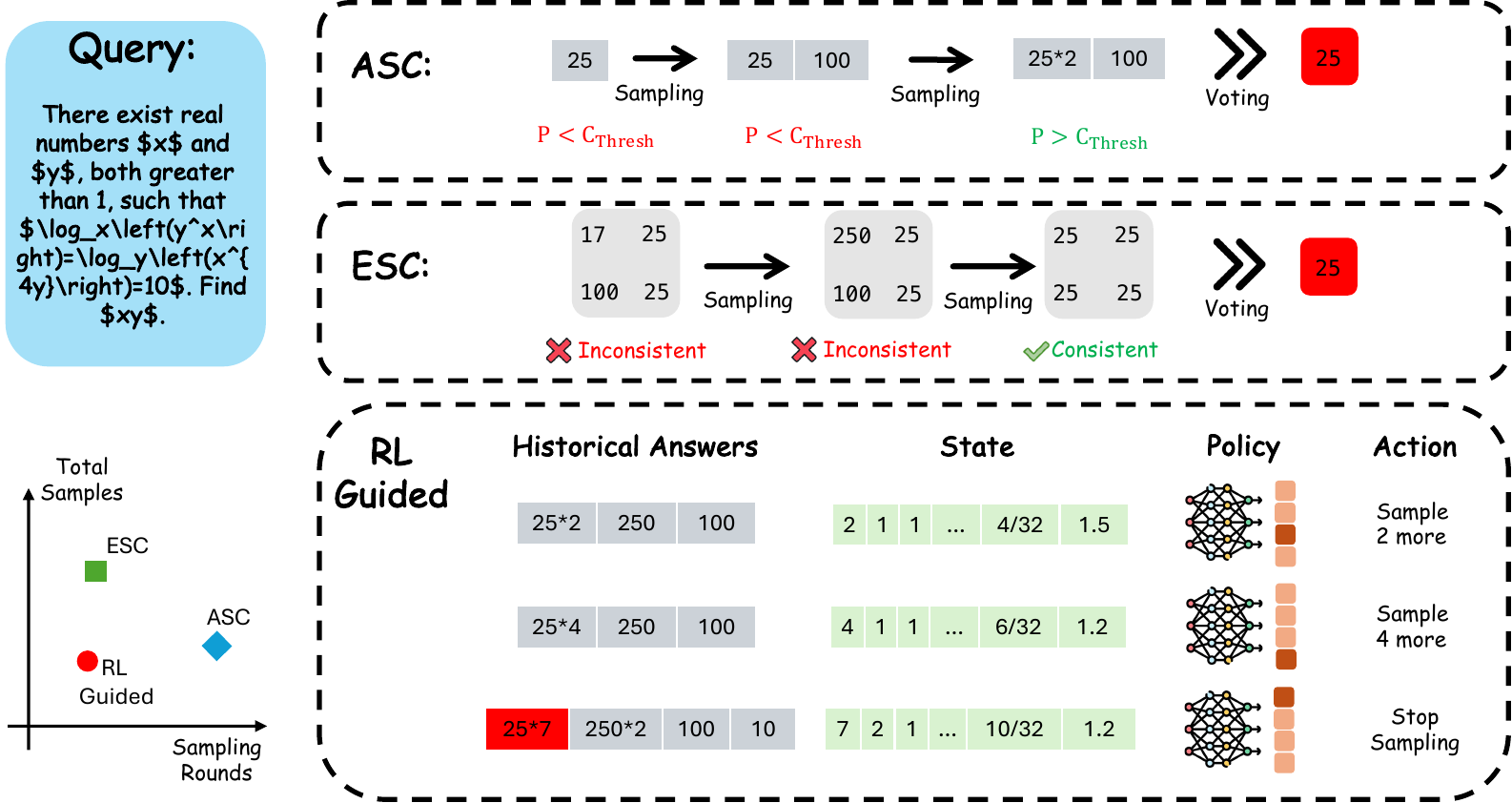}
  \caption{\textbf{Overview of the \method{} framework.} Top two blocks illustrates mechanism of two adaptive sampling baselines. ASC sequentially samples one response at a time and stops when the posterior probability exceeds a predefined threshold. ESC samples in fixed batches and stops only when intra-batch consistency is achieved. Different from those approaches, \method{} guides the sampling process via a lightweight policy network. At each round, the framework constructs a state based on statistics from the observed answer pool. Given this state, the policy network directs the language model to either sample a specific number of additional responses in parallel or halt generation. As summarized in the bottom-left plot, \method{} outperforms baseline methods by requiring fewer total samples and sampling rounds."}
  \label{fig:main_figure}
\end{figure*}

While existing approaches are effective, they often suffer from several key limitations: (i) Many rely heavily on human-designed heuristics or distributional assumptions, rather than explicitly deriving an optimal policy to navigate the performance-cost trade-off. (ii) Several methods require auxiliary signals that are often unavailable under specific scenarios, such as internal model confidence, hidden states, or question difficulty. (iii) Some techniques are highly invasive, disrupting the model's natural reasoning process or even requiring additional training of the underlying LLM sampler. Such modifications incur significant operational overhead and are often incompatible with standard inference pipelines. These limitations motivate the need for \textit{a lightweight but principled method for efficient test-time scaling.} Furthermore, to align with practical use cases, the method should flexibly account for multiple objectives, such as latency and computational cost.

In this work, we present a framework that circumvents these limitations by formulating adaptive sampling as a Markov decision process (MDP). Rather than relying on predefined rules, we train a four-layer MLP controller via RL to learn a policy that optimizes the performance-cost trade-off. Crucially, our method relies purely on statistics derived from the sampled answer set, requiring only the final generated answers. This means it demands no auxiliary features, such as model confidence, and requires no intervention in the reasoning process of the LLMs. In the environment, we jointly consider multiple objectives. We use the correctness of the final answer as a positive reward, while treating inference costs (additional samples and sampling rounds) as penalties. Given an observed answer set, the controller dynamically decides whether to sample a specific number of additional responses or to stop sampling, as illustrated in Figure \ref{fig:main_figure}.



Practically, \method{} is extremely lightweight. The controller itself is only a small four-layer MLP, which can be trained and deployed efficiently on CPUs. Theoretically, our formulation intuitively maps to a constrained optimization problem: maximizing answer accuracy under strict latency and computation budgets. The weighted objective of our RL controller naturally arises as the Lagrangian relaxation of this problem, providing a principled mechanism to trade off performance and cost.

We validate \method{} across three benchmarks and multiple language-model samplers. Our experiments show that the proposed controller consistently improves the accuracy--efficiency trade-off over strong adaptive sampling baselines. Specifically, as shown in Table \ref{tab:main_table}, \method{} reduces sampling rounds and total samples by 3 times and 30\%, respectively, compared to ASC \citep{aggarwal2023let}, and by 10\% and 35\% compared to ESC \citep{li2024escape}. This improvement is consistent across different trade-off levels (Figure \ref{fig:4b-count}). Furthermore, the learned policy generalizes well: controllers trained using samples from one dataset or model can be transferred to different benchmarks and even different samplers with minimal performance degradation (Section \ref{sec:generalization}).

\section{Method}
\subsection{Problem Setup and MDP Formulation}
We formulate the adaptive sampling problem as a finite-horizon Markov decision process (MDP). At each round, the controller observes the statistical features of current pool of sampled answers, and then decides whether to stop or to acquire some additional samples. The reward is designed to reflect three factors: the quality of the final aggregated answer, the latency cost induced by additional sampling rounds, and the computation cost induced by generating more candidates. 

Consider an input query $x$. Let $N$ be the maximum number of samples allowed. At any given sampling round $t$, let $\mathcal{D}_t = \{y_1, \dots, y_{n_t}\}$ denote the observed answer set, where each $y_i \in \mathcal{Y}$ is the final answer extracted from the $i$-th response, e.g., the content inside \verb|\boxed{}|. Here, $n_t$ denotes the total number of candidates collected up to round $t$.

\paragraph{State.}
The state $s_t$ summarizes the evidence currently available to the controller, together with the inference cost already incurred. Specifically, the state consists of three components: the counts of the most frequent answer classes, the total number of sampled candidates and some light statistics. Let
\[
\mathcal{V}_t = \{v_1(t), v_2(t), \dots, v_K(t)\}
\]
denote the sorted counts of the top-$K$ most frequent answers in $\mathcal{D}_t$. We then represent the state using $\mathcal{V}_t$, the total number of samples generated $|\mathcal{D}_t|$, and the entropy of $\mathcal{V}_t$, such that $s_t = \{{\mathcal{V}_t, |n_t|, \text{Ent}(\mathcal{V}_t)} \}$.

\paragraph{Action and transition.}
At each round, the controller selects an action from
\[
\mathcal{A} = \{0, k_1, k_2, \dots, k_L\}.
\]
The action $a_t=0$ means stopping the sampling process and returning the current majority-vote answer. An action $a_t=k_\ell>0$ means generating $k_\ell$ additional candidate answers in parallel. 

Following the generation of new samples, the observed answer set $\mathcal{D}_t$ is updated, and the state $\mathcal{V}_t$ is recomputed before advancing to the subsequent round. The episode terminates either when the controller chooses to stop (i.e., $a_t=0$) or when the maximum sampling budget is reached (i.e., $n_t+a_t \geq N$).

\paragraph{Reward Function}

The reward function is designed to balance the final answer's quality against the inference cost. We can explicitly divide this into two distinct components: the step-wise penalty and the terminal reward.

\textit{Step-wise Penalty.}
At each round $t$, if the controller chooses to continue sampling by selecting a non-zero action $a_t > 0$, it incurs an intermediate penalty:
\begin{equation*}
    r_{t}^{\text{step}} = -\lambda_{\mathrm{lat}} - \lambda_{\mathrm{comp}} a_t,
\end{equation*}
where $\lambda_{\mathrm{lat}}, \lambda_{\mathrm{comp}} \geq 0$. Essentially, this penalizes the model by $\lambda_{\mathrm{lat}}$ for taking a new sampling step and by $\lambda_{\mathrm{comp}}$ for each new sample generated.

\textit{Terminal Reward.}
The episode terminates either when the controller chooses to stop or when the maximum sampling budget is reached. Upon termination, the controller aggregates the current candidate pool $\mathcal{D}_t$ via majority voting to output a final prediction:
\begin{equation*}
    \hat{y}_t = \mathrm{MajorityVote}(\mathcal{D}_t).
\end{equation*}
The model then receives a terminal reward $r^{\text{final}}_t$ based on the correctness of the prediction:
\begin{equation*}
    r^{\text{final}}_t =
    \begin{cases}
    1, & \text{if } \hat{y}_t = y^\star, \\
    -1, & \text{otherwise}.
    \end{cases}
\end{equation*}

Specifically, we define $y^\star$ as the majority-vote answer that would be obtained if sampling continued to the maximum budget $N$ without early stopping. This target encourages the controller to halt generation as soon as the current answer distribution converges. Notably, this reward design is intentionally decoupled from question-specific signals, such as ground-truth labels. This maintains consistency with our state representation, which depends solely on the sampled answer pool. We further validate these reward design choices through an ablation study in Section~\ref{sec:ablation}.

By combining these two components, the reward $r_t$ at any sampling step $t$ can be formally expressed as follows:
\begin{equation*}
    r_t = \mathbb{I}(a_t > 0)r_t^{\text{step}} + \mathbb{I}(a_t = 0 \text{ or } n_t+a_t \geq N) r^{\text{final}}_t
\end{equation*}

\paragraph{Optimization Objective.}
Given the per-step reward $r_t$, our goal is to train the controller to maximize the expected cumulative reward. Let $t_{\mathrm{stop}}$ denote the actual step at which the episode terminates. The optimization objective for the policy $\pi_\theta$ can be concisely formulated as:
\begin{equation}
    J(\pi_\theta) = \mathbb{E}_{\pi_\theta} \left[ \sum_{t=0}^{t_{\mathrm{stop}}} r_t \right],
    \label{eq:target}
\end{equation}
where the expectation is taken over the sequence of states and actions induced by $\pi_\theta$.

The proposed environment is compatible with a broad class of RL algorithms. In this work, we adopt Proximal Policy Optimization (PPO) as our default training algorithm, as it provides a stable on-policy framework for optimizing stochastic policies. However, alternative approaches, such as value-based methods (e.g., DQN) or other policy-gradient algorithms, can be readily applied within the same MDP formulation.

Furthermore, existing sampling strategies, such as ASC and ESC, can be naturally interpreted within this framework as fixed, rule-based policies operating over the proposed MDP. In contrast, our PPO-based controller dynamically learns an optimal strategy by explicitly maximizing the expected cumulative reward $J(\pi_\theta)$.

\subsection{Lagrangian View}
\label{sec:lagrangian_cpo}

The optimization objective in Eq.~\eqref{eq:target} admits a simple Lagrangian interpretation. Intuitively, adaptive sampling can be viewed as a resource-constrained optimization problem: the controller seeks to maximize final answer quality while satisfying constraints on expected latency and computation. The penalties on sampling rounds and generated candidates then arise naturally as dual variables associated with these resource constraints.

\begin{proposition}[Lagrangian interpretation]
\label{prop:lagrangian}
Let $J_{\mathrm{ans}}(\pi_\theta)$, $J_{\mathrm{comp}}(\pi_\theta)$, and
$J_{\mathrm{lat}}(\pi_\theta)$ denote the expected accuracy, total number of samples,
and number of sampling rounds under policy $\pi_\theta$, respectively. Consider the
budget-constrained adaptive sampling problem
\begin{equation*}
\begin{aligned}
\max_{\pi} \quad & J_{\mathrm{ans}}(\pi_\theta) \\
\mathrm{s.t.} \quad
& J_{\mathrm{comp}}(\pi_\theta) \leq C_{\mathrm{comp}}, \\
& J_{\mathrm{lat}}(\pi_\theta) \leq C_{\mathrm{lat}} .
\end{aligned}
\end{equation*}
Optimizing the objective in Equation~\ref{eq:target} is equivalent to optimizing
the Lagrangian relaxation of this constrained problem with respect to $\pi$,
where the non-negative penalty weights act as dual variables associated with
the sample and round constraints.
\end{proposition}

We provide the formal statement and proof in Appendix~\ref{app:lagrangian}. Proposition~\ref{prop:lagrangian} provides an interpretation of our reward design from a constrained optimization perspective. This connects \method{} to a broader literature on constrained and safe RL \citep{garcia2015comprehensive, altman2021constrained}, where policies are optimized under explicit cost or safety constraints. While we use fixed penalty in this work, this perspective suggests a promising future direction: directly optimizing policies under prescribed budgets with constrained RL methods. 

\section{Experimental Settings}
\subsection{Dataset and Samplers}

In this paper, we evaluate our method on questions from three challenging mathematical reasoning benchmarks: AIME24, AIME25 \citep{aime}, and HMMT 2025~\citep{dekoninck2026matharena}. This benchmark selection aligns with prior work~\cite{zheng2026parallel}, which is intended to provide a balanced level of difficulty.  To train the RL-guided sampling controller, we randomly sample a 200-question subset from the DAPO training set \citep{yu2025dapo}.

We further evaluate the scalability and generalizability of RL-guided sampling across a diverse set of LLM samplers. Specifically, we consider multiple variants of the Qwen-3 family~\citep{yang2025qwen3} and the closed-source GPT-4.1-nano model~\citep{achiam2023gpt}. These samplers span different model scales (0.6B, 1.7B, and 4B), model types (reasoning and instruct), and deployment settings (open-source and proprietary). This broad selection allows us to examine whether the benefits of RL-guided sampling transfer consistently from lightweight open-source models to more capable proprietary systems.

\subsection{Baseline Methods and Evaluation Metrics}

To evaluate the effectiveness of RL-Guided Sampling, we compare it against representative test-time scaling baselines:
\begin{itemize}[leftmargin=*]
    \item \textbf{SC} (Self-Consistency~\cite{wang2022self}): A standard test-time scaling method that samples multiple independent reasoning trajectories in parallel and returns the majority-voted answer.
    
    \item \textbf{ASC} (Adaptive Self-Consistency~\cite{aggarwal2023let}): An adaptive sampling method that sequentially samples one response at a time, updates the posterior distribution, and stops once a predefined p-value threshold is reached.
    
    \item \textbf{ESC} (Early Stopping Consistency~\cite{li2024escape}): A chunk-based approach that generates a fixed number of trajectories in parallel at each step and terminates early when the responses within a batch are consistent.
\end{itemize}

We report performance using five key metrics, grouped into three categories. 
First, \textbf{Accuracy} measures the percentage of correctly solved problems. 
Second, \textbf{Total Samples} and \textbf{Total Tokens} measure computational cost: the former is directly aligned with our environment design, while the latter reflects the actual token cost during deployment. 
Third, \textbf{Sampling Rounds} and \textbf{Sequential Tokens} measure latency: the former captures the number of adaptive sampling steps, while the latter measures the latency-critical sequential token length in real-time inference.

\subsection{Environment and Training Setup}
For state construction, we set $K=5$, resulting in an 7-dimensional state space. The action space $\mathcal{A} = \{0,1,2,4\}$ additional responses. The policy network is a four-layer MLP. Full implementation details of the RL training procedure and environment setup are provided in Appendix~\ref{sec:detail}. 

\section{ Results and Analysis}

\begin{table*}[h!]
\centering
\begin{adjustbox}{width=\textwidth}
\begin{tabular}{l ccc ccc ccc ccc}
\toprule
Method & \multicolumn{3}{c}{AIME24} & \multicolumn{3}{c}{AIME25} & \multicolumn{3}{c}{HMMT25} & \multicolumn{3}{c}{Avg.} \\
\cmidrule(lr){2-4}\cmidrule(lr){5-7}\cmidrule(lr){8-10}\cmidrule(lr){11-13}
& Acc. $\uparrow$ & Rounds $\downarrow$ & \#Samples $\downarrow$ & Acc. $\uparrow$ & Rounds $\downarrow$ & \#Samples $\downarrow$ & Acc. $\uparrow$ & Rounds $\downarrow$ & \#Samples $\downarrow$ & Acc. $\uparrow$ & Rounds $\downarrow$ & \#Samples $\downarrow$ \\
\midrule
\multicolumn{13}{l}{\textit{Model: Qwen3-0.6B-Thinking}} \\
\midrule
SC@32 & 22.6 & 1.0 & 32.0 & 30.1 & 1.0 & 32.0 & 16.4 & 1.0 & 32.0 & 23.0 & 1.0 & 32.0 \\
ASC & 22.6 & 27.1 & 27.1 & 30.1 & 24.2 & 24.2 & 16.4 & 23.2 & 23.2 & 23.0 & 24.8 \textsuperscript{\scriptsize(\textcolor{green!80!black}{+2383.3\%})} & 24.8 \textsuperscript{\scriptsize(\textcolor{red}{-22.4\%})} \\
ESC & 22.3 & 6.1 & 30.7 & 29.8 & 5.8 & 28.9 & 16.4 & 5.9 & 29.7 & 22.9 & 5.9 \textsuperscript{\scriptsize(\textcolor{green!80!black}{+493.3\%})} & 29.8 \textsuperscript{\scriptsize(\textcolor{red}{-7.0\%})} \\
RL-guided & 22.6$_{\pm0.06}$ & 5.5 & 21.9 & 30.0$_{\pm0.04}$ & 5.0 & 19.9 & 16.4$_{\pm0.02}$ & 4.5 & 18.0 & 23.0 & 5.0 \textsuperscript{\scriptsize(\textcolor{green!80!black}{+400.0\%})} & 19.9 \textsuperscript{\scriptsize(\textcolor{red}{-37.7\%})} \\
\midrule
\multicolumn{13}{l}{\textit{Model: Qwen3-1.7B-Thinking}} \\
\midrule
SC@32 & 68.3 & 1.0 & 32.0 & 44.1 & 1.0 & 32.0 & 25.9 & 1.0 & 32.0 & 46.1 & 1.0 & 32.0 \\
ASC & 68.2 & 17.7 & 17.7 & 44.2 & 18.7 & 18.7 & 26.0 & 20.2 & 20.2 & 46.1 & 18.9 \textsuperscript{\scriptsize(\textcolor{green!80!black}{+1786.7\%})} & 18.9 \textsuperscript{\scriptsize(\textcolor{red}{-41.0\%})} \\
ESC & 67.3 & 4.7 & 23.6 & 44.1 & 4.6 & 22.8 & 26.4 & 4.9 & 24.6 & 45.9 & 4.7 \textsuperscript{\scriptsize(\textcolor{green!80!black}{+373.3\%})} & 23.7 \textsuperscript{\scriptsize(\textcolor{red}{-26.0\%})} \\
RL-guided & 67.6$_{\pm0.04}$ & 2.9 & 11.0 & 44.6$_{\pm0.02}$ & 3.5 & 14.0 & 26.7$_{\pm0.03}$ & 3.5 & 14.0 & 46.3 & 3.3 \textsuperscript{\scriptsize(\textcolor{green!80!black}{+230.0\%})} & 13.0 \textsuperscript{\scriptsize(\textcolor{red}{-59.4\%})} \\
\midrule
\multicolumn{13}{l}{\textit{Model: Qwen3-4B-Instruct-Thinking}} \\
\midrule
SC@32 & 73.3 & 1.0 & 32.0 & 57.5 & 1.0 & 32.0 & 33.6 & 1.0 & 32.0 & 54.8 & 1.0 & 32.0 \\
ASC & 73.3 & 13.8 & 13.8 & 57.5 & 15.0 & 15.0 & 33.6 & 17.0 & 17.0 & 54.8 & 15.3 \textsuperscript{\scriptsize(\textcolor{green!80!black}{+1426.7\%})} & 15.3 \textsuperscript{\scriptsize(\textcolor{red}{-52.3\%})} \\
ESC & 72.7 & 3.6 & 18.0 & 57.0 & 4.2 & 20.8 & 33.6 & 4.4 & 22.0 & 54.4 & 4.1 \textsuperscript{\scriptsize(\textcolor{green!80!black}{+306.7\%})} & 20.3 \textsuperscript{\scriptsize(\textcolor{red}{-36.7\%})} \\
RL-guided & 73.0$_{\pm0.07}$ & 2.5 & 10.0 & 57.1$_{\pm0.14}$ & 2.8 & 11.0 & 33.6$_{\pm0.03}$ & 3.0 & 11.8 & 54.6 & 2.8 \textsuperscript{\scriptsize(\textcolor{green!80!black}{+176.7\%})} & 10.9 \textsuperscript{\scriptsize(\textcolor{red}{-65.8\%})} \\
\midrule
\multicolumn{13}{l}{\textit{Model: GPT-4.1-nano}} \\
\midrule
SC@32 & 37.1 & 1.0 & 32.0 & 33.5 & 1.0 & 32.0 & 12.7 & 1.0 & 32.0 & 27.8 & 1.0 & 32.0 \\
ASC & 37.1 & 20.6 & 20.6 & 33.5 & 20.8 & 20.8 & 12.7 & 23.7 & 23.7 & 27.8 & 21.7 \textsuperscript{\scriptsize(\textcolor{green!80!black}{+2070.0\%})} & 21.7 \textsuperscript{\scriptsize(\textcolor{red}{-32.2\%})} \\
ESC & 36.9 & 5.2 & 25.8 & 33.2 & 5.2 & 26.0 & 12.3 & 5.6 & 28.2 & 27.5 & 5.3 \textsuperscript{\scriptsize(\textcolor{green!80!black}{+433.3\%})} & 26.7 \textsuperscript{\scriptsize(\textcolor{red}{-16.7\%})} \\
RL-guided & 36.9$_{\pm0.04}$ & 6.5 & 16.4 & 33.4$_{\pm0.05}$ & 6.9 & 17.3 & 12.6$_{\pm0.02}$ & 7.3 & 18.6 & 27.7 & 6.9 \textsuperscript{\scriptsize(\textcolor{green!80!black}{+590.0\%})} & 17.4 \textsuperscript{\scriptsize(\textcolor{red}{-45.5\%})} \\
\midrule
\end{tabular}
\end{adjustbox}
\caption{
\textbf{Comparison of test-time scaling approaches across three benchmarks.}
Acc.\ denotes accuracy, Rounds measures the number of adaptive sampling rounds, and \# Samples counts the total number of sampled responses. In general, better methods achieve higher Acc.\ with fewer Rounds and \# Samples. For the Self-Consistency baseline, we use 32 samples, denoted as SC@32. For ASC and ESC, we follow the default settings from the original papers, setting the ASC threshold to $0.95$ and the ESC chunk size to 5. For RL-guided adapters, we report the mean performance over five random seeds, together with the standard deviation.}
\label{tab:main_table}
\end{table*}

\subsection{Main Results}
\label{sec:main_results}
The main results are reported in Table~\ref{tab:main_table} and Table~\ref{tab:main_token_table} in the Appendix. The former reports performance in terms of sampling rounds and total samples, while the latter provides token-level metrics, including total tokens and sequential tokens. 

Overall, \method{} consistently achieves a better accuracy--efficiency trade-off than strong baselines. The key observations are as follows:

\begin{itemize}[leftmargin=*]
    \item ASC effectively reduces total sample consumption but suffers from high latency due to its reliance on excessive sequential sampling rounds. In contrast, ESC evaluates on a coarse grid to reduce these rounds, but it compromises performance by either requiring more total samples or sacrificing accuracy.

    \item Compared to ASC, \method{} significantly alleviates the latency bottleneck by reducing average sampling rounds by $3$--$4\times$ while preserving comparable accuracy. Furthermore, it achieves this efficiency while reducing the total number of samples by approximately $30\%$.

    \item Compared to ESC, \method{} avoids overly aggressive early stopping to achieve a better overall trade-off. It slightly reduces the number of sampling rounds by approximately $10\%$, while significantly lowering the total sample requirement by roughly $33\%$ and achieving higher accuracy.

    \item Compared to SC, all adaptive methods (ASC, ESC, and \method{}) significantly reduce total sample count. Nevertheless, their sampling rounds inevitably exceed those of SC due to the sequential evaluations required for adaptive early stopping.
\end{itemize}

\subsection{Scaling Curves}
\label{sec:scaling}

\begin{figure}[h!]
\centering
\includegraphics[width=0.48\textwidth]{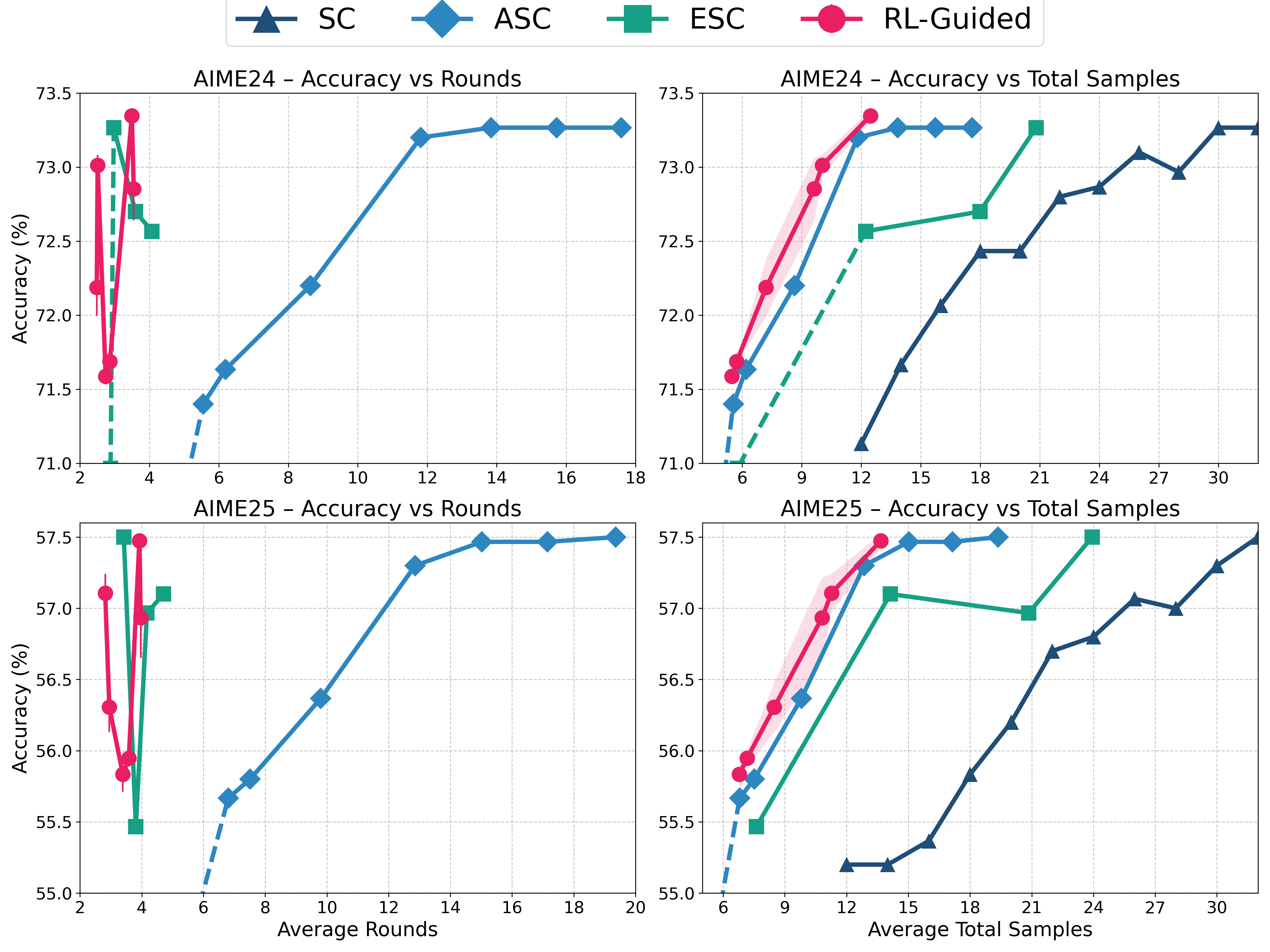}
\caption{Accuracy scaling behavior across sampling budgets. Left: Accuracy vs. Sampling Rounds. Right: Accuracy vs. Total Samples. Results are generated with Qwen3-4B-Instruct on the AIME24 and AIME25 datasets. Compared to SC, ASC, and ESC, RL-Guided sampling consistently achieves superior accuracy under the same or fewer samples and rounds. Analogous scaling curves measured by token consumption are presented in Figure~\ref{fig:4b-seq}.}
\label{fig:4b-count}
\end{figure}

Figure~\ref{fig:4b-count} illustrates the test-time scaling behavior of RL-Guided Sampling under various hyperparameter settings, comparing it against SC, ASC, and ESC using Qwen3-4B-Instruct on the AIME 2024 and 2025 datasets. Each curve represents a parameter sweep, capturing different preferences along the accuracy--efficiency trade-off. Detailed experimental setups and additional scaling results for other models and datasets are provided in Appendix~\ref{sec:appendix_scaling}.

Overall, these results align with the findings in Section~\ref{sec:main_results}, demonstrating that \method{} consistently achieves a favorable trade-off from both efficiency perspectives. In terms of total sample budget, RL-Guided slightly outperforms ASC and significantly dominates both ESC and SC across most operating points. Furthermore, when evaluated by sampling rounds, both RL-Guided and ESC maintain strong performance while requiring significantly fewer rounds. Ultimately, these results validate the effectiveness and stability of \method{} across a broad spectrum of parameter configurations, demonstrating robust performance well beyond default settings.

\subsection{Explanatory Analysis}
\label{sec:explain}

\begin{figure}[h!]
\centering
\includegraphics[width=0.48\textwidth]{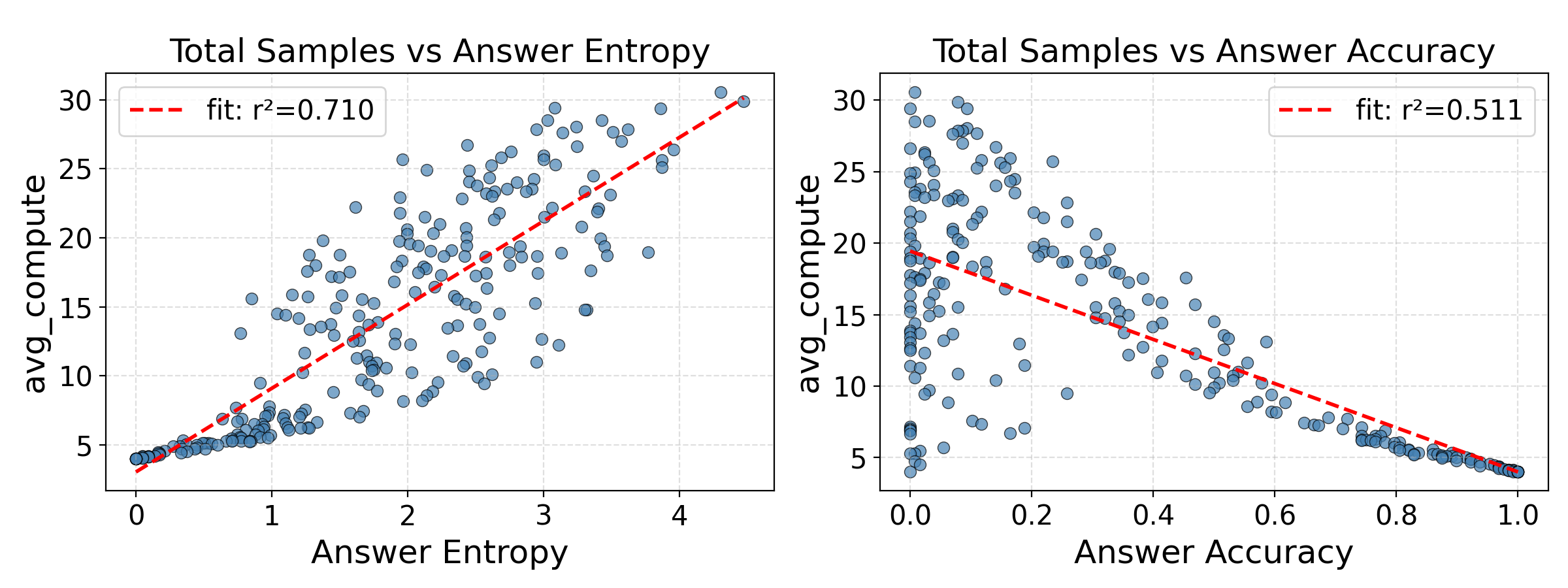}
\caption{Correlation between the average total samples per query and Answer Entropy (left) alongside Answer Accuracy (right). Each data point represents a distinct query from the DAPO-subset, with responses generated by the Qwen3-0.6B model.}
\label{fig:interpret}
\end{figure}

Compared to Self-Consistency (SC) with a fixed sample size, the primary advantage of adaptive sampling lies in its ability to dynamically allocate computational resources across queries. In this section, we investigate how \method{} distributes this computation. Specifically, we record the average total samples consumed per query and examine its correlation with two established query-level metrics, as illustrated in Figure~\ref{fig:interpret}.

Given a specific query, we define the following two metrics: \textbf{Answer Entropy} is the Shannon entropy of the categorical distribution over the final answers, treating each unique generated answer as a distinct category. \textbf{Answer Accuracy} denotes the empirical probability of generating a correct response, calculated over the total sampled responses.

\begin{table*}[t!]
\centering
\begin{adjustbox}{width=\textwidth}
\begin{tabular}{l ccc ccc ccc ccc}
\toprule
Reward Signal & \multicolumn{3}{c}{AIME24} & \multicolumn{3}{c}{AIME25} & \multicolumn{3}{c}{HMMT25} & \multicolumn{3}{c}{Avg.} \\
\cmidrule(lr){2-4}\cmidrule(lr){5-7}\cmidrule(lr){8-10}\cmidrule(lr){11-13}
& Acc. $\uparrow$ & Rounds $\downarrow$ & \#Samples $\downarrow$ & Acc. $\uparrow$ & Rounds $\downarrow$ & \#Samples $\downarrow$ & Acc. $\uparrow$ & Rounds $\downarrow$ & \#Samples $\downarrow$ & Acc. $\uparrow$ & Rounds $\downarrow$ & \#Samples $\downarrow$ \\
\midrule
Running Majority & 73.0$_{\pm0.07}$ & 2.5 & 10.0 & 57.1$_{\pm0.14}$ & 2.8 & 11.0 & 33.6$_{\pm0.03}$ & 3.0 & 11.8 & 54.6 & 2.8 & 10.9 \\
Full Majority & 73.0$_{\pm0.11}$ & 2.6 & 11.0 & 57.1$_{\pm0.19}$ & 2.9 & 12.0 & 33.6$_{\pm0.04}$ & 3.0 & 12.2 & 54.6 & 2.8 \textsuperscript{\scriptsize(\textcolor{green!80!black}{+2.4\%})} & 11.7 \textsuperscript{\scriptsize(\textcolor{green!80!black}{+7.3\%})} \\
Real Label & 71.9$_{\pm0.27}$ & 3.7 & 15.0 & 55.7$_{\pm0.39}$ & 4.3 & 18.0 & 33.6$_{\pm0.12}$ & 4.3 & 18.0 & 53.7 & 4.1 \textsuperscript{\scriptsize(\textcolor{green!80!black}{+48.2\%})} & 17.0 \textsuperscript{\scriptsize(\textcolor{green!80!black}{+55.9\%})} \\
\midrule
\end{tabular}
\end{adjustbox}
\caption{Ablation study evaluating the impact of different reward signals across three reasoning benchmarks using Qwen3-4B-Instruct. We compare our default \textit{Running Majority} target against a \textit{Full Majority} target and the ground-truth \textit{Real Label}. The \textit{Running Majority} configuration achieves the best overall performance. Notably, utilizing the \textit{Real Label} as the reward signal results in a substantial degradation in accuracy, accompanied by significant increase in both sampling rounds and total generated samples. All experimental settings are identical to those in Table~\ref{tab:main_table}.}
\label{tab:reward_ablation}
\end{table*}

The results demonstrate that RL-Guided sampling learns an intuitive resource allocation strategy. Specifically, as illustrated in Figure~\ref{fig:interpret}, the average sample budget allocated by the policy exhibits a distinct positive correlation with Answer Entropy. This aligns with the intuition that queries generating highly diverse answers inherently require more samples to converge and form a reliable majority. Notably, the resource allocation does not strictly mirror the entropy score, suggesting that the learned policy captures the nuanced dynamics of the sampling trajectory rather than applying a rigid heuristic threshold. In contrast, the correlation with Answer Accuracy is comparatively weak. This aligns with our formulation, as the RL controller relies purely on the real-time statistics of the ongoing sampling process without incorporating query-level information.

\subsection{Generalization Analysis}
\begin{figure}[h!]
\centering
\includegraphics[width=0.48\textwidth]{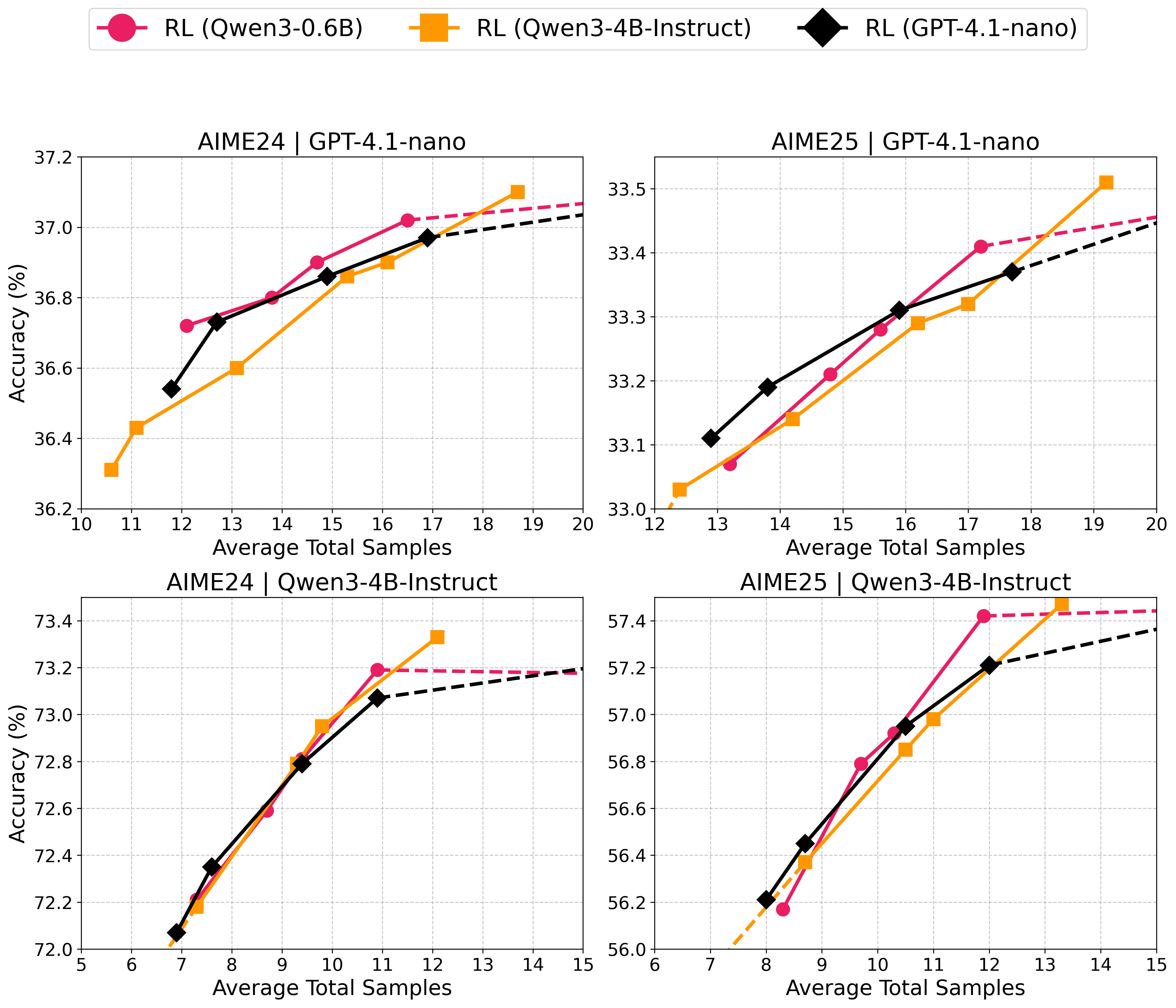}
\caption{Accuracy scaling behavior of \method{} evaluated on GPT-4.1-nano (top) and Qwen3-4B-Instruct (bottom). Lines denote controllers trained on the DAPO-subset using responses generated by different models, demonstrating robust cross-model transferability.}
\label{fig:generalization}
\end{figure}

Having previously validated that \method{} generalizes across different datasets using the same model in section \ref{sec:main_results} and \ref{sec:scaling}, we now evaluate a more challenging setting: generalizing across \textit{both} models and datasets simultaneously. Specifically, Figure~\ref{fig:generalization} illustrates the performance of a controller trained on the Qwen3-0.6B model when applied directly to guide inference for the closed-source GPT-4.1-nano model. As shown, the learned policy demonstrates strong robustness, maintaining highly competitive scaling behavior across various parameter settings despite the distribution shift.
\label{sec:generalization}

This strong generalizability carries significant practical implications. It suggests that the controller successfully captures model-agnostic signals of sampling convergence rather than overfitting to a specific model's distribution. Consequently, practitioners can train the policy using existing data or inexpensive samples from a lightweight open-source model, and seamlessly apply it to reduce inference costs for new, stronger, or closed-source models where direct sampling is costly or infeasible.

\subsection{Ablation Study}
\label{sec:ablation}

We investigate the impact of different reward signals by comparing three target label ($y^\star$) constructions for the terminal reward: Running Majority (default; the majority vote if sampling continues to the $N=32$ budget, Full Majority (the majority vote from the full answer pool of $128$ samples), and Real Label (the ground-truth answer). The results are summarized in Table \ref{tab:reward_ablation}.

As shown, the Running Majority target yields optimal accuracy-efficiency trade-off. Conversely, utilizing the Real Label significantly degrades accuracy and increases sample cost. This underperformance relates to our state representation. Because the state excludes the semantic information of the query, the controller must base its decisions on the statistical distribution of the sampled answer pool. Consequently, the controller lacks the context to verify factual accuracy. Using the Real Label therefore introduces indistinguishable noise into the optimization process. Furthermore, incorporating such problem-dependent signals prevents the policy from learning universal stopping criteria, ultimately degrading its generalization performance.

Furthermore, Running Majority slightly outperforms Full Majority in sampling efficiency. We hypothesize that the actual samples limit ($N=32$) provides a more attainable positive reward signal compared to forcing the policy to predict the consensus of a much larger sample pool.

\section{Related works}
\label{sec:related}
\subsection{Efficient Parallel Reasoning}
A recent line of work seeks to reduce the cost of fixed-budget parallel sampling through dynamic resource allocation. \citet{aggarwal2023let} and \citet{li2024escape} terminate sampling once a consensus criterion is reached, and \citet{wang2025make} further adapt the sample budget to query difficulty. A complementary direction weights reasoning paths by confidence to recover high-quality answers from fewer samples~\citep{huang2025efficient,Taubenfeld2025ConfidenceIS,Fu2025DeepTW}. These methods, however, largely rely on sequential sampling, which undermines the hardware advantage of parallel decoding. Finer-grained schemes such as Dynamic Self-Consistency~\citep{wan2025reasoning}, Self-Truncation~\citep{wang2025sampling}, DeepPrune~\cite{tu2025deepprune}, Step~\cite{liang2026hidden}, Slim-SC~\cite{hong2025slim}, Parallel-Probe~\cite{zheng2026parallel} instead prune unpromising trajectories during generation to avoid wasted compute on incorrect paths. More recently, AutoTTS~\cite{zheng2026llms} explores letting LLMs themselves discover better test-time scaling strategies. Yet these methods rely on intrinsic signals (e.g., logits) or mid-generation rollouts to decide when to prune, complicating deployment. We instead cast parallel test-time scaling as an MDP and search for a lightweight controller that jointly optimizes accuracy, latency, and cost, yielding better Pareto trade-offs.


\subsection{Test-Time Scaling}
Improving the efficiency of complex reasoning has increasingly been framed as a question of how to allocate test-time computation~\citep{snell2024scaling, chen2025iterative,wang2026time}. A prominent instantiation is tree search, which aggregates diverse reasoning paths and uses sparse activation to keep the search tractable~\citep{bi2024forest, lample2022hypertreeproofsearchneural, koh2024treesearchlanguagemodel, zheng2025parallel}. Step-wise verifiers further tighten this search by pruning unproductive branches on the fly~\citep{wang2022self, li2023makinglargelanguagemodels, lightman2023letsverifystepstep}. Orthogonal to search itself, gains have been reported from diversifying query formulations~\citep{Huang2024DivideRA} and from iterative refinement loops that bootstrap the model's self-correction on harder problems~\citep{chen2025sets, welleck2022generatingsequenceslearningselfcorrect, madaan2023selfrefineiterativerefinementselffeedback, aggarwal2024alphaverusbootstrappingformallyverified,wang2026not}.

\subsection{Multi-Objective and Constrained Reinforcement Learning}
Multi-Objective Reinforcement Learning (MORL) addresses sequential decision-making problems governed by conflicting goals, such as maximizing diagnostic yield while minimizing invasive tests in healthcare~\citep{qiu2026optimizing}. By optimizing across multiple reward signals, MORL seeks to discover a set of Pareto-optimal policies that capture different trade-offs~\citep{roijers2013survey, hayes2022practical}. A foundational approach in MORL is linear scalarization, which aggregates the multi-dimensional objective into a single scalar reward using pre-defined preference weights ~\citep{parisi2014policy, mossalam2016multi}. Alternatively, conditioned approaches incorporate the preference vector directly into the state space, aiming to learn a single unified policy $\pi(a|s, w)$ that generalizes across the entire Pareto front~\citep{abels2019dynamic,yang2019generalized,navon2020learning}. Given its algorithmic simplicity and robust empirical performance, our proposed method relies on the linear scalarization paradigm.

Constrained Reinforcement Learning (CRL) focuses on maximizing a primary objective while satisfying several limits, such as safety boundaries or resource budgets~\citep{garcia2015comprehensive, altman2021constrained}. Through Lagrangian relaxation, this constrained formulation is fundamentally connected to MORL. Mainstream CRL methods employ dynamic dual updates or bounded optimizations to strictly enforce these constraints (e.g., Constrained Policy Optimization~\citep{achiam2017constrained} and primal-dual approaches~\citep{tessler2018reward}).

\section{Conclusion}
This paper introduces \method{}, a principled and lightweight framework for efficient test-time scaling. By formulating adaptive sampling as a Markov decision process, we train an RL controller to explicitly optimize the trade-off among answer correctness, computational cost, and latency. Unlike existing approaches, \method{} is non-invasive and relies purely on the statistics of the generated answers, eliminating the need for auxiliary signals. Empirically, our CPU-friendly controller consistently outperforms strong baselines such as ASC and ESC, significantly reducing both total samples and sampling rounds. Furthermore, the learned policy exhibits strong transferability across different datasets and samplers.


\section*{Limitations}
This work takes an initial step toward formulating adaptive LLM sampling as a reinforcement learning problem. While our lightweight controller already shows consistent improvements in the accuracy--latency--compute trade-off, the current formulation can be further refined. 

In particular, our state representation intentionally relies on simple statistics of the sampled answers, leaving room to incorporate richer signals such as answer confidence, or the average length of answers. More importantly, on reward design, future work could directly include real-life costs, such as directly using time and money cost of the generating process as penalty. These extensions are complementary to our framework and may further improve its alignment with real-world deployment costs.
\bibliography{custom}


\appendix

\section{Implementing Details}
\label{sec:detail}
For each question, we generate 128 candidate responses. The sampling configuration largely follows the recommended parameters for each model, as summarized in Table~\ref{tab:generating-detail}. During both training and evaluation, each trajectory is constructed by randomly sampling $N=32$ responses from this candidate pool. We use the following prompt:
\begin{promptbox}
\{Query\}

Please reason step by step, and put your final answer within \texttt{\textbackslash boxed\{\}}.
\end{promptbox}

\begin{table}[h!]
\centering
\resizebox{\columnwidth}{!}{
\begin{tabular}{lcccc}
\toprule
\textbf{Model} & \textbf{Max Length} & \textbf{Temperature} & \textbf{Top-$k$} & \textbf{Top-$p$} \\
\midrule
Qwen3-0.6B-Thinking & 32768 & 0.6 & 20 & 0.95 \\
Qwen3-1.7B-Thinking & 32768 & 0.6 & 20 & 0.95 \\
Qwen3-4B-Instruct   & 32768 & 0.7 & 20 & 0.8 \\
GPT-4.1-nano        & 32768 & 0.8 & 20 & 0.95 \\
\bottomrule
\end{tabular}}
\caption{Sampling configurations used to generate candidate responses for each model.}
\label{tab:generating-detail}
\end{table}

The RL adapter is a four layer neural network. We train the RL guided adapter with the PPO method implemented with the Stable-Baselines3 framework \citep{stable-baselines3} and the MDP environment is built following the OpenAI gym framework \citep{1606.01540}. Detailed training configurations for both methods are provided in Table~\ref{tab:training-detail}. The experiments are conducted on a linux platform with 64 INTEL(R) XEON(R) GOLD 6548Y+ CPU.

During training, we first randomly select questions from the DAPO subset. For each selected question, we construct a trajectory by randomly sampling a batch $N=32$ responses from its offline response pool of 128 candidates, and use the majority-vote answer within this batch as the pseudo-label $y^*$. During evaluation, we iterate over all questions in each benchmark. For each question, we repeatedly sample 32 responses from the corresponding pool of 128 candidates and evaluate all methods under the same sampled response sets. We repeat this process over 100 random seeds and report the mean accuracy and computation cost.

\begin{table}[h!]
\centering
\begin{tabular}{lc}
\toprule
\textbf{Parameter} & \textbf{Value} \\
\midrule
learning\_rate & $1\times10^{-5}$ \\
total\_timesteps & $1\times10^{6}$ \\
gamma (Discount Factor) & 1 \\
gae\_lambda  & 0.95 \\
Hidden layers & [32, 64, 64, 32] \\
\bottomrule
\end{tabular}
\caption{Training hyperparameter of \method{}.}
\label{tab:training-detail}
\end{table}

\section{Additional Experimental Results}
\subsection{Additional Details and Results for Main Results}
\label{appendix:main}
For the results in Table~\ref{tab:main_table}, we use a default penalty setting of $\lambda_{\mathrm{lat}}=0$ and $\lambda_{\mathrm{comp}}=0.0075$ when training the RL-guided controller for all models. For each sampler, the controller is trained on the DAPO subset using responses generated by that same sampler. To account for randomness in RL training, we train five controllers with different random seeds for each model. We report the mean sampling rounds, total number of samples, and accuracy across these runs, with the standard deviation reported for accuracy.

In addition to Table~\ref{tab:main_table}, we report the performance of \method{} and competing baselines using token-level cost metrics in Table~\ref{tab:main_token_table}. Specifically, we measure computation cost by Total Tokens and latency by Sequential Tokens. The results are consistent with those in Section~\ref{sec:main_results}, further confirming the accuracy--efficiency trade-off achieved by \method{}.

\begin{table*}[ht!]
\centering
\begin{adjustbox}{width=\textwidth}
\begin{tabular}{l ccc ccc ccc ccc}
\toprule
Method & \multicolumn{3}{c}{AIME24} & \multicolumn{3}{c}{AIME25} & \multicolumn{3}{c}{HMMT25} & \multicolumn{3}{c}{Avg.} \\
\cmidrule(lr){2-4}\cmidrule(lr){5-7}\cmidrule(lr){8-10}\cmidrule(lr){11-13}
& Acc. $\uparrow$ & Seq. Tokens $\downarrow$ & ToT. Tokens $\downarrow$ & Acc. $\uparrow$ & Seq. Tokens $\downarrow$ & ToT. Tokens $\downarrow$ & Acc. $\uparrow$ & Seq. Tokens $\downarrow$ & ToT. Tokens $\downarrow$ & Acc. $\uparrow$ & Seq. Tokens $\downarrow$ & ToT. Tokens $\downarrow$ \\
\midrule

\multicolumn{13}{l}{\textit{Model: Qwen3-0.6B-Thinking}} \\
\midrule
SC@32 & 22.6 & 24.7k & 495.6k & 30.1 & 23.5k & 441.1k & 16.4 & 23.8k & 458.8k & 23.0 & 24.0k & 465.2k \\
ASC & 22.6 & 437.1k & 437.1k & 30.1 & 359.5k & 359.5k & 16.4 & 349.4k & 349.4k & 23.0 & 382.0k \textsuperscript{\scriptsize(\textcolor{green!80!black}{+1492.0\%})} & 382.0k \textsuperscript{\scriptsize(\textcolor{red}{-17.9\%})} \\
ESC & 22.3 & 138.9k & 480.3k & 29.8 & 121.0k & 415.2k & 16.4 & 127.9k & 436.9k & 22.9 & 129.3k \textsuperscript{\scriptsize(\textcolor{green!80!black}{+438.7\%})} & 444.1k \textsuperscript{\scriptsize(\textcolor{red}{-4.5\%})} \\
RL-guided & 22.6$_{\pm0.06}$ & 115.7k & 356.0k & 30.0$_{\pm0.04}$ & 97.3k & 298.7k & 16.4$_{\pm0.02}$ & 89.8k & 272.7k & 23.0 & 100.9k \textsuperscript{\scriptsize(\textcolor{green!80!black}{+320.6\%})} & 309.1k \textsuperscript{\scriptsize(\textcolor{red}{-33.6\%})} \\
\midrule
\multicolumn{13}{l}{\textit{Model: Qwen3-1.7B-Thinking}} \\
\midrule
SC@32 & 68.3 & 26.2k & 494.1k & 44.1 & 25.8k & 511.5k & 25.9 & 27.7k & 587.8k & 46.1 & 26.6k & 531.1k \\
ASC & 68.2 & 317.7k & 317.7k & 44.2 & 358.8k & 358.8k & 26.0 & 420.6k & 420.6k & 46.1 & 365.7k \textsuperscript{\scriptsize(\textcolor{green!80!black}{+1275.0\%})} & 365.7k \textsuperscript{\scriptsize(\textcolor{red}{-31.1\%})} \\
ESC & 67.3 & 118.7k & 412.8k & 44.1 & 116.6k & 421.7k & 26.4 & 135.4k & 505.8k & 45.9 & 123.6k \textsuperscript{\scriptsize(\textcolor{green!80!black}{+364.6\%})} & 446.8k \textsuperscript{\scriptsize(\textcolor{red}{-15.9\%})} \\
RL-guided & 67.6$_{\pm0.04}$ & 67.6k & 210.2k & 44.6$_{\pm0.02}$ & 83.9k & 271.6k & 26.7$_{\pm0.03}$ & 90.0k & 296.0k & 46.3 & 80.5k \textsuperscript{\scriptsize(\textcolor{green!80!black}{+202.7\%})} & 259.2k \textsuperscript{\scriptsize(\textcolor{red}{-51.2\%})} \\
\midrule
\multicolumn{13}{l}{\textit{Model: Qwen3-4B-Instruct-Thinking}} \\
\midrule
SC@32 & 73.3 & 13.0k & 221.3k & 57.5 & 13.1k & 215.3k & 33.6 & 15.0k & 244.1k & 54.8 & 13.7k & 226.9k \\
ASC & 73.3 & 127.8k & 127.8k & 57.5 & 126.6k & 126.6k & 33.6 & 147.1k & 147.1k & 54.8 & 133.8k \textsuperscript{\scriptsize(\textcolor{green!80!black}{+875.8\%})} & 133.8k \textsuperscript{\scriptsize(\textcolor{red}{-41.0\%})} \\
ESC & 72.7 & 48.2k & 161.5k & 57.0 & 49.7k & 167.3k & 33.6 & 57.1k & 189.7k & 54.4 & 51.7k \textsuperscript{\scriptsize(\textcolor{green!80!black}{+277.0\%})} & 172.8k \textsuperscript{\scriptsize(\textcolor{red}{-23.8\%})} \\
RL-guided & 73.0$_{\pm0.07}$ & 30.8k & 90.1k & 57.1$_{\pm0.14}$ & 32.1k & 96.2k & 33.6$_{\pm0.03}$ & 34.6k & 101.2k & 54.6 & 32.5k \textsuperscript{\scriptsize(\textcolor{green!80!black}{+136.7\%})} & 95.8k \textsuperscript{\scriptsize(\textcolor{red}{-57.8\%})} \\
\midrule
\multicolumn{13}{l}{\textit{Model: GPT-4.1-nano}} \\
\midrule
SC@32 & 37.1 & 5.3k & 92.7k & 33.5 & 5.0k & 86.5k & 12.7 & 4.2k & 80.7k & 27.8 & 4.8k & 86.6k \\
ASC & 37.1 & 68.6k & 68.6k & 33.5 & 59.7k & 59.7k & 12.7 & 62.5k & 62.5k & 27.8 & 63.6k \textsuperscript{\scriptsize(\textcolor{green!80!black}{+1219.0\%})} & 63.6k \textsuperscript{\scriptsize(\textcolor{red}{-26.6\%})} \\
ESC & 36.9 & 24.1k & 81.1k & 33.2 & 22.4k & 75.2k & 12.3 & 20.9k & 73.3k & 27.5 & 22.5k \textsuperscript{\scriptsize(\textcolor{green!80!black}{+365.7\%})} & 76.5k \textsuperscript{\scriptsize(\textcolor{red}{-11.7\%})} \\
RL-guided & 36.9$_{\pm0.04}$ & 26.0k & 55.4k & 33.4$_{\pm0.05}$ & 23.0k & 49.2k & 12.6$_{\pm0.02}$ & 22.8k & 49.9k & 27.7 & 24.0k \textsuperscript{\scriptsize(\textcolor{green!80!black}{+396.7\%})} & 51.5k \textsuperscript{\scriptsize(\textcolor{red}{-40.6\%})} \\
\midrule
\end{tabular}
\end{adjustbox}
\caption{
\textbf{Comparison of test-time scaling approaches across three benchmarks.}
Acc.\ denotes accuracy, Seq.\ Tokens measures Sequential Tokens, and Tot.\ Tokens measures Total Tokens. We use the same parameter setup as in Table~\ref{tab:main_table}.}
\label{tab:main_token_table}
\end{table*}

\begin{figure}[h!]
\centering
\includegraphics[width=0.49\textwidth]{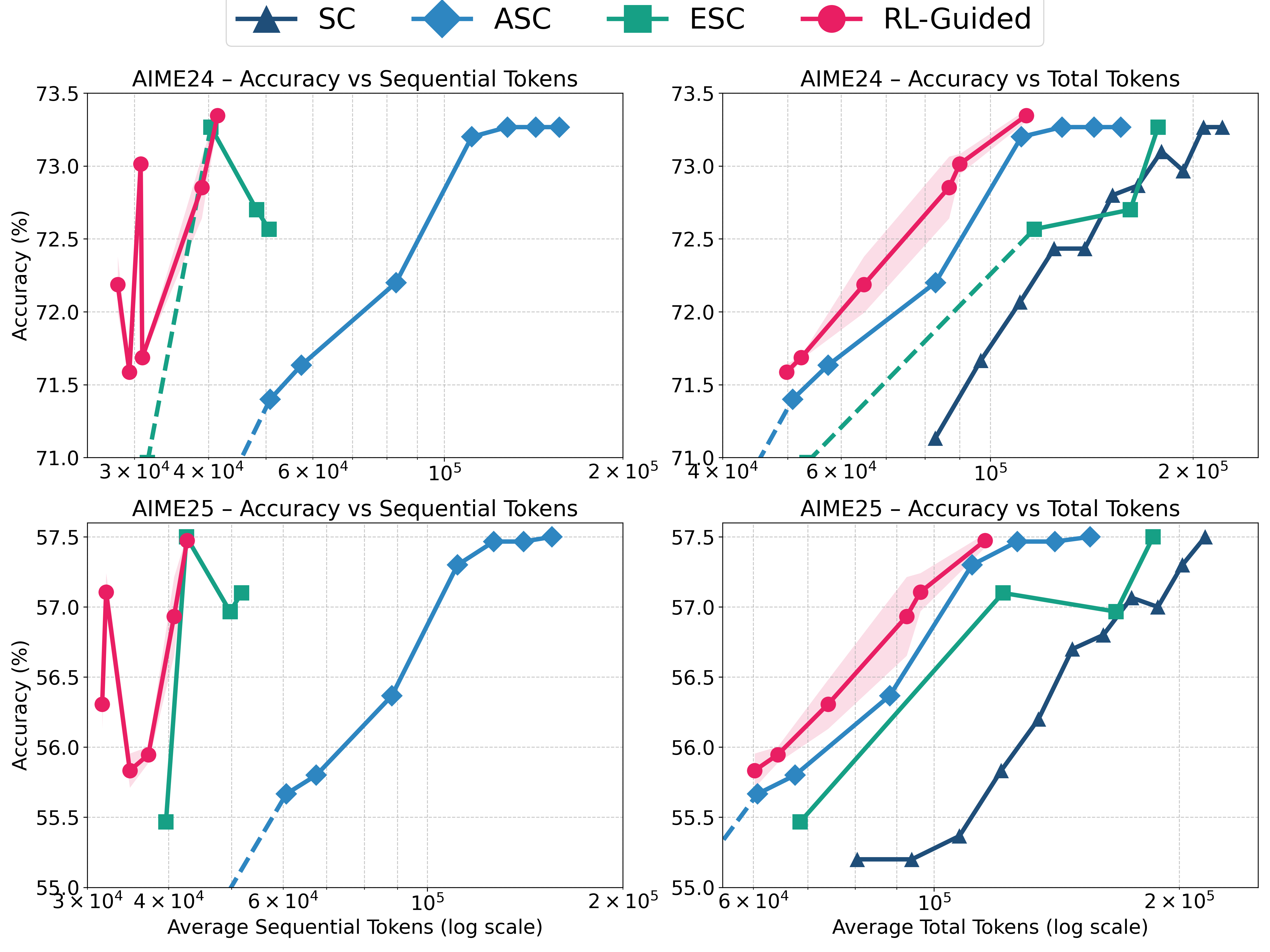}
\caption{Accuracy--token scaling curves comparing the SC, ASC, ESC and \method{}. across different models and benchmarks. Results are generated with Qwen3-4B-Instruct on the AIME24 and AIME25 datasets.}
\label{fig:4b-seq}
\end{figure}

\subsection{Additional Details and Results for Scaling Analysis}\label{sec:appendix_scaling}
\begin{figure*}[h!]
  \centering
  \includegraphics[width=\linewidth]{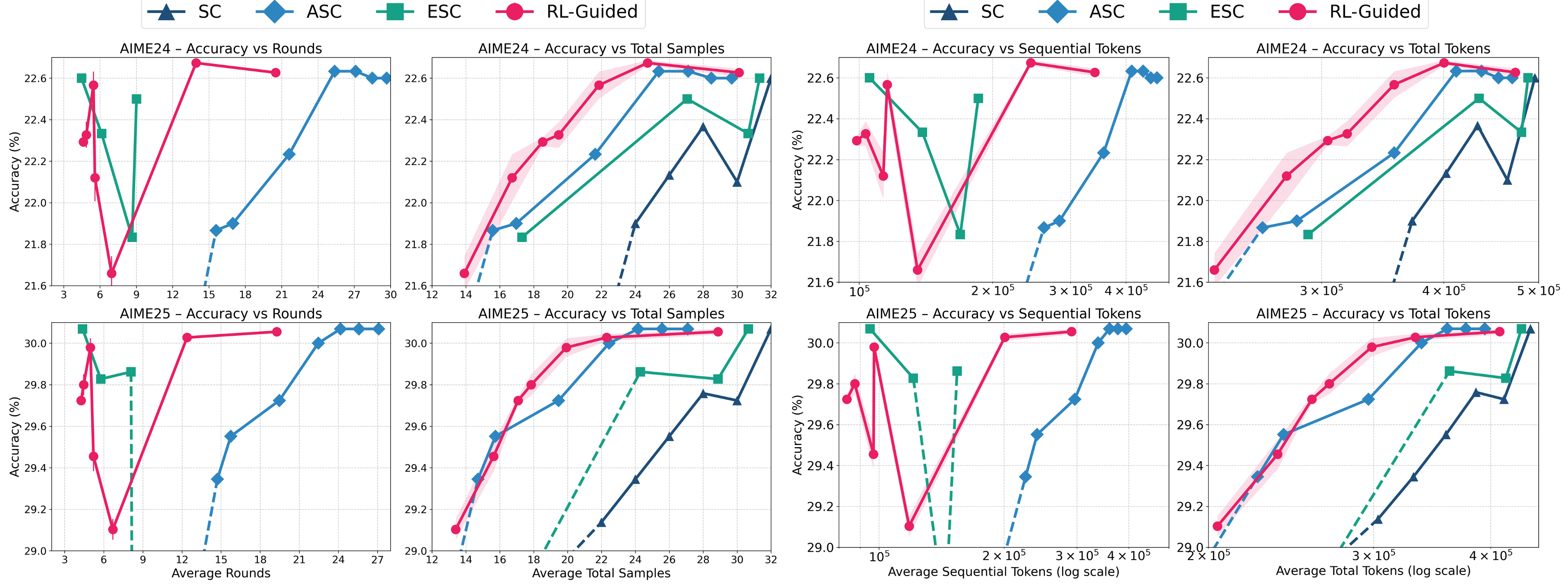}
  \caption{Accuracy--token scaling curves(right) and Accuracy--sampling scaling curves(left) comparing the SC, ASC, ESC and \method{}. across different models and benchmarks. Results are generated with Qwen3-0.6B on the AIME24 and AIME25 datasets.}
  \label{fig:0.6-combined}
\end{figure*}

For Figure~\ref{fig:4b-count}, we conduct a parameter sweep for \method{} with 
$\lambda_{\mathrm{lat}}=0$ and 
$\lambda_{\mathrm{comp}} \in \{0.001, 0.005, 0.0075, 0.015, 0.02, 0.03\}$. 
For ASC, we sweep the stopping threshold 
$C_{\text{threshold}} \in \{0.6, 0.65, 0.75, 0.8, 0.87, 0.92, 0.95, 0.98, 0.99\}$, 
and for ESC, we sweep the chunk size 
$K \in \{2,3,5,7\}$. 
In addition to sample-level metrics, Figure~\ref{fig:4b-seq} reports the corresponding token-level scaling curves, where Total Tokens measure computation cost and Sequential Tokens measure latency. 
We further provide scaling curves using Qwen3-0.6B as the LLM sampler in Figure~\ref{fig:0.6-combined}.


\subsection{Additional Results for Explanatory Analysis}

\begin{figure}[h!]
\centering
\includegraphics[width=0.48\textwidth]{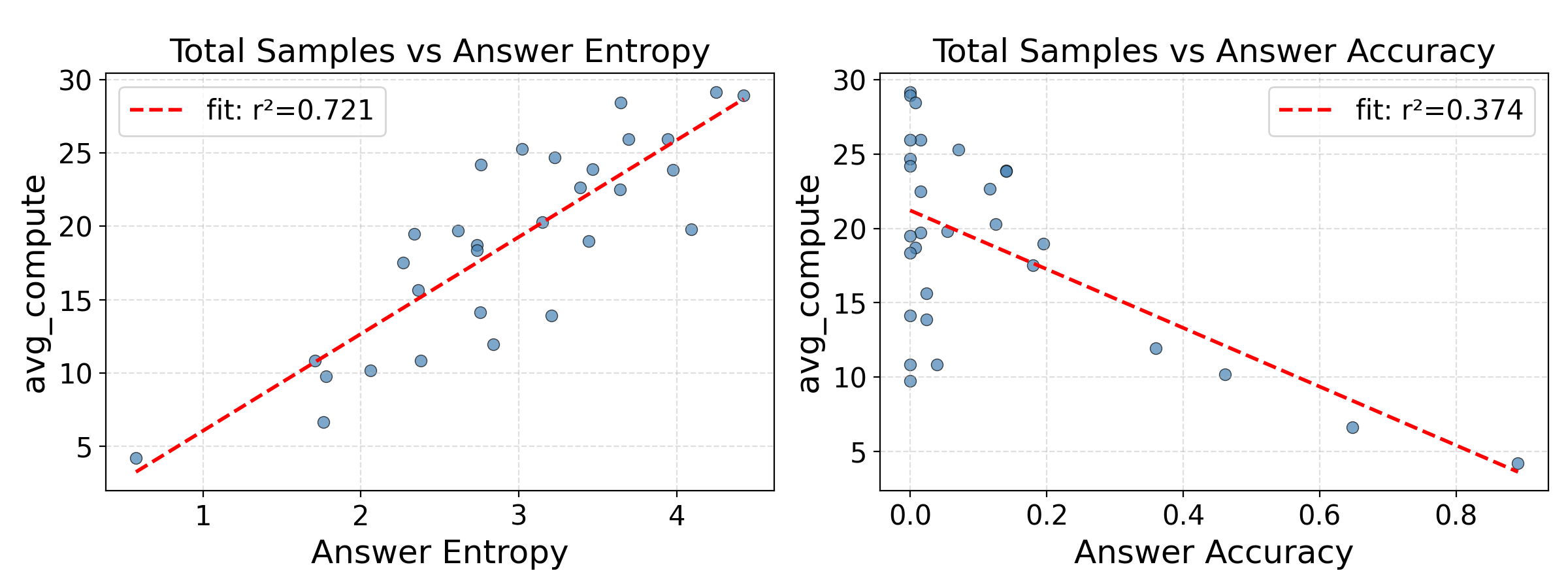}
\caption{Correlation between total samples per query and Answer Entropy (left) alongside Answer Accuracy (right). Each point represents a distinct query from AIME24, with responses generated by Qwen3-0.6B.}
\label{fig:AIME-explain}
\end{figure}

In Section~\ref{sec:explain}, we present the explanatory analysis on the DAPO subset, which contains more questions and therefore provides clearer visualization. For completeness, Figure~\ref{fig:AIME-explain} reports the corresponding analysis on the AIME24 dataset, where we observe a similar trend.

\clearpage
\onecolumn
\raggedbottom

\section{Proof of Proposition~\ref{prop:lagrangian}}
\label{app:lagrangian}

This section provides the theoretical justification for Proposition~\ref{prop:lagrangian}, demonstrating the equivalence between our reinforcement learning objective and the Lagrangian relaxation of the budget-constrained adaptive sampling problem. 

First, we clarify the definition of expected accuracy $J_{\mathrm{ans}}(\pi_\theta)$ in the context of our MDP. Because our environment is intentionally decoupled from ground-truth labels, the terminal reward evaluates the final prediction against the maximum-budget consensus $y^\star$. Thus, $J_{\mathrm{ans}}(\pi_\theta)$ mathematically represents the expected agreement with this asymptotic majority vote, serving as our proxy for accuracy. Furthermore, because $r^{\mathrm{final}} \in \{1, -1\}$, its expectation is an affine transformation of the raw matching probability $P(\hat{y} = y^\star)$, specifically $2P - 1$. We define $J_{\mathrm{ans}}(\pi_\theta) = \mathbb{E}_{\tau \sim \pi_\theta} \left[ r^{\mathrm{final}}_{t_{\mathrm{stop}}} \right]$ to encapsulate this scaled objective.

In our proposed MDP, the controller's objective is to maximize the expected episodic return $J(\pi_\theta) = \mathbb{E}_{\tau \sim \pi_\theta}[R(\tau)]$. Let $t_{\mathrm{stop}}$ denote the additional sampling steps taken before the episode terminates. The total return is the sum of rewards accumulated across all steps, $R(\tau) = \sum_{t=0}^{t_{\mathrm{stop}}} r_t$. 

By separating the terminal reward $r^{\mathrm{final}}_{t_{\mathrm{stop}}}$ from the intermediate step penalties incurred during generation, we can expand the expected return as follows:
$$
\begin{aligned}
J(\pi_\theta) &= \mathbb{E}_{\tau \sim \pi_\theta} \left[ r^{\mathrm{final}}_{t_{\mathrm{stop}}} + \sum_{t=0}^{t_{\mathrm{stop}}-1} (-\lambda_{\mathrm{lat}} - \lambda_{\mathrm{comp}} a_t) \right] \\
&= \mathbb{E}_{\tau \sim \pi_\theta} \left[ r^{\mathrm{final}}_{t_{\mathrm{stop}}} \right] - \lambda_{\mathrm{lat}} \mathbb{E}_{\tau \sim \pi_\theta} [t_{\mathrm{stop}}] - \lambda_{\mathrm{comp}} \mathbb{E}_{\tau \sim \pi_\theta} \left[ \sum_{t=0}^{t_{\mathrm{stop}}-1} a_t \right].
\end{aligned}
$$

To strictly map these terms to the total metrics defined in Proposition~\ref{prop:lagrangian}, we must account for the initial observation round $t_0$ and the initial candidate pool size $n_0$ used to construct the initial state $s_0$. The expected total latency (sampling rounds) is $J_{\mathrm{lat}}(\pi_\theta) = t_0 + \mathbb{E}_{\tau \sim \pi_\theta} [t_{\mathrm{stop}}]$, and the expected total computation (generated samples) is $J_{\mathrm{comp}}(\pi_\theta) = n_0 + \mathbb{E}_{\tau \sim \pi_\theta} \left[ \sum_{t=0}^{t_{\mathrm{stop}}-1} a_t \right]$. 

Substituting these relations back into the expanded return equation yields:
$$
J(\pi_\theta) = J_{\mathrm{ans}}(\pi_\theta) - \lambda_{\mathrm{lat}} \big(J_{\mathrm{lat}}(\pi_\theta) - t_0\big) - \lambda_{\mathrm{comp}} \big(J_{\mathrm{comp}}(\pi_\theta) - n_0\big).
$$

Now, consider the budget-constrained adaptive sampling problem defined in Proposition~\ref{prop:lagrangian}, which seeks to maximize final expected accuracy proxy subject to predefined operational budgets:
$$
\begin{aligned}
\max_{\pi_\theta} \quad & J_{\mathrm{ans}}(\pi_\theta) \\
\mathrm{s.t.} \quad & J_{\mathrm{lat}}(\pi_\theta) - C_{\mathrm{lat}} \leq 0, \\
& J_{\mathrm{comp}}(\pi_\theta) - C_{\mathrm{comp}} \leq 0.
\end{aligned}
$$

By introducing non-negative dual variables $\lambda_{\mathrm{lat}}, \lambda_{\mathrm{comp}} \ge 0$ for the inequality constraints, the Lagrangian relaxation of this problem is:
$$
\mathcal{L}(\pi_\theta, \lambda_{\mathrm{lat}}, \lambda_{\mathrm{comp}}) = J_{\mathrm{ans}}(\pi_\theta) - \lambda_{\mathrm{lat}} \big(J_{\mathrm{lat}}(\pi_\theta) - C_{\mathrm{lat}}\big) - \lambda_{\mathrm{comp}} \big(J_{\mathrm{comp}}(\pi_\theta) - C_{\mathrm{comp}}\big).
$$

Rearranging the Lagrangian to separate the policy-dependent components from the constant terms gives:
$$
\mathcal{L}(\pi_\theta, \lambda_{\mathrm{lat}}, \lambda_{\mathrm{comp}}) = J_{\mathrm{ans}}(\pi_\theta) - \lambda_{\mathrm{lat}} J_{\mathrm{lat}}(\pi_\theta) - \lambda_{\mathrm{comp}} J_{\mathrm{comp}}(\pi_\theta) + \lambda_{\mathrm{lat}} C_{\mathrm{lat}} + \lambda_{\mathrm{comp}} C_{\mathrm{comp}}.
$$

Comparing this expression with our derived $J(\pi_\theta)$, we establish the exact algebraic relationship:
$$
\mathcal{L}(\pi_\theta, \lambda_{\mathrm{lat}}, \lambda_{\mathrm{comp}}) = J(\pi_\theta) - \lambda_{\mathrm{lat}} t_0 - \lambda_{\mathrm{comp}} n_0 + \lambda_{\mathrm{lat}} C_{\mathrm{lat}} + \lambda_{\mathrm{comp}} C_{\mathrm{comp}}.
$$

Because the scaling mappings and the offset terms (involving $t_0$, $n_0$, $C_{\mathrm{lat}}$, and $C_{\mathrm{comp}}$) are strictly affine transformations independent of the policy $\pi_\theta$, maximizing the Lagrangian $\mathcal{L}$ with respect to $\pi_\theta$ is mathematically equivalent to maximizing our unconstrained RL objective $J(\pi_\theta)$. This explicitly confirms that optimizing the step-wise rewards in our defined MDP solves the Lagrangian relaxation of the constrained adaptive sampling problem. \hfill $\blacksquare$

\end{document}